\documentclass[letterpaper, 10 pt, conference]{ieeeconf}  

\IEEEoverridecommandlockouts                              
\def\ie{{i.e.}}
\def\eg{{e.g.}}
\newcommand{\C}{$\mathcal{C}$}
\newcommand{\W}{$\mathcal{W}$}

\usepackage{amsmath} 
\usepackage{graphicx}
\usepackage{amssymb}
\usepackage[latin9]{inputenc}
\usepackage{verbatim}
\usepackage[english]{babel}
\usepackage{subcaption}
\usepackage[table,xcdraw]{xcolor}
\usepackage[obeyFinal,colorinlistoftodos,textsize=small]{todonotes}   
\usepackage{booktabs}
\title{\LARGE \bf
	Near-optimal Smooth Path Planning for Multisection Continuum Arms
}

\author{Jiahao Deng, Brandon H. Meng, Iyad Kanj, and Isuru S. Godage
	\thanks{The authors are affiliated with the School of Computing, DePaul University, Chicago, IL 60604, USA. $\{\mbox{jdeng5,bmeng1,ikanj,igodage}\}$@depaul.edu. \newline\newline This work was supported in part by the National Science Foundation (NSF) grant number~1718755 and DePaul University Academic Initiative Pool grant number~601709.}
}

\begin{document}

\maketitle
\thispagestyle{empty}
\pagestyle{empty}

\begin{abstract}
	We study the path planning problem for continuum-arm robots, in which we are given a starting and an end point, and we need to compute a path for the tip of the continuum arm between the two points. We consider both cases where obstacles are present and where they are not. 
	
	We demonstrate how to leverage the continuum arm features to introduce a new model that enables a path planning approach based on the configurations graph, for a continuum arm consisting of three sections, each consisting of three muscle actuators.  The algorithm we apply to the configurations graph allows us to exploit parallelism in the computation to obtain efficient implementation. 
	
	We conducted extensive tests, and the obtained results show the completeness of the proposed algorithm under the considered discretizations, in both cases where obstacles are present and where they are not. We compared our approach to the standard inverse kinematics approach. While the inverse kinematics approach is much faster when successful, our algorithm always succeeds in finding a path or reporting that no path exists, compared to a roughly 70\% success rate of the inverse kinematics approach (when a path exists).
	
\end{abstract}

\section{INTRODUCTION}\label{sec:intro}
In this paper, we study the path planning problem for continuum-arm robots. In the path planing problem, we are given two points, a starting point and a destination/end point, and we need to compute a path for the tip of the continuum arm from the starting point to the destination point. We consider both cases where obstacles are present and where they are not. 

Continuum arms, such as trunk and tentacle robots, attempt to mimic the biological appendages---like elephant trunks---and have seen a surge of research in recent years~\cite{bio,webster,palmer2019active,santiago2016soft}.
These robots lie between the two extremities of rigid and soft robots, and promise to capture the ``best'' of both worlds in terms of manipulability/elasticity, degrees of freedom (DoF), and compliance. Their need spans industrial, manufacturing, and healthcare applications, in which safe human-robot collaboration is essential~\cite{haddadin,sanan,zinn,robinson}. 
In spite of their demonstrated capabilities in manipulability, superior performance in constrained spaces, and compliant operation, their full potential has not been realized yet. This lag is largely due to their inherent physical features, which, while making continuum arms human-friendly, contribute to complexities in 
path planning and control. 

A continuum arm consists usually of multiple sections, each consisting of several pneumatic muscle actuators \cite{mcmahan2006field,godage2016dynamics}. 
Figure~\ref{fig:scenario} shows two pneumatically actuated multisection continuum arms prototypes assuming smooth bending deformation in each section to achieve complex poses.
Due to the redundancy in the number of DoF they possess, the mapping between their configuration space (\C-Space) and their task space (\W-space) is nonlinear, which makes path planning for such robots a complex problem. (This redundancy is present since multiple \C-space configurations of the continuum arm correspond to the same point in its \W-space.) As a result, little work has been done on path planning for continuum arms~\cite{godage2012path,ataka2016real}. 
The prevailing mathematical tool used for path planning relies on inverse kinematics (IK). Due to the high redundancy alluded to above, IK-based methods are not reliable, and result in a high failure rate \cite{godage2015modal,godage2015dual}.
These methods have the potential of getting stuck in local minima, especially if obstacles are present in the \W-space (such as in the application depicted in Fig.~\ref{fig:scenario}). Indeed, it has been shown in~\cite{godage2015dual} that IK methods can be unreliable. 
Although they have not been applied yet to continuum manipulators, sampling-based path planning methods present another possibility for performing path planning (see~\cite{survey}).
However, we mention that path planning approaches that are based on sampling  the \C-space (\eg, PRM and RRT) suffer from the lack of performance guarantee, and are not suitable for sensitive applications of continuum-arm robots (\eg, healthcare applications) \cite{burgner2015continuum}.  

\begin{center}
	\begin{figure}[t]
		\begin{centering}
			\includegraphics[width=1.0\columnwidth]{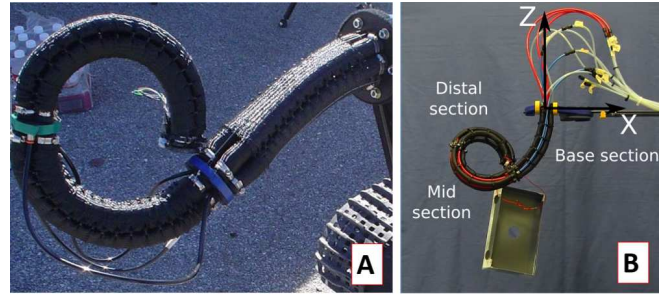}
			
			\par\end{centering}
		\caption{Pneumatically actuated multisection continuum arms: (A) OctArm-IV \cite{mcmahan2006field}; and (B) a continuum arm, designed at the Italian Institute of Technology, handling an object \cite{godage2015modal}.}
		\label{fig:scenario}
	\end{figure}
	\par
\end{center} 
\vspace*{-6mm}

A reliable approach, \ie, one that provides performance guarantee, for performing path planning is based on the \C-space graph, which captures the whole range of motion of the entire continuum arm, even in the presence of obstacles. In such approach, one chooses proper discretizations of the \W-space and the \C-Space, and constructs a mapping from the \C-space to the \W-space w.r.t.~these discretizations. A graph is then constructed, whose vertices are the \C-space configurations and whose edges represent the adjacencies between configurations w.r.t.~a 1-step variance in the discretization of the \C-space. Path planning is then performed on this graph, and mapped to the \W-space. In general, the issue with the above approach is that it becomes infeasible for systems with high DoF, as the number of configurations is exponential.  

In this paper, we demonstrate how to leverage the continuum arm features to introduce a new model that enables a path planning approach based on the \C-space graph, for a continuum arm consisting of three sections, each consisting of three muscle actuators bundled together. This model allows discretizations of the \W-space and \C-space to a high-fidelity degree, resulting in a reliable and smooth path planning algorithm. The graph algorithm we apply to the \C-space graph allows us to exploit parallelism in the computation to obtain efficient implementation. We conducted extensive tests. The obtained results show the completeness of the proposed algorithm under the considered discretizations, in both cases where obstacles are present and where they are not: The algorithm provides a path between the initial and final point in the \W-space when such a path exists, and otherwise, reports that no path exists. Moreover, the algorithm does so within a reasonable amount of time. We compared our approach to the standard IK approach. While the IK approach is much faster when successful, our algorithm always succeeds in finding a path or reporting that no path exists, compared to a roughly 70\% success rate of the IK approach (when a path exists).

\section{SYSTEM MODEL}\label{sec:System-Model}
\begin{figure}[tb]
	\begin{centering}
		\includegraphics[width=1.0\columnwidth]{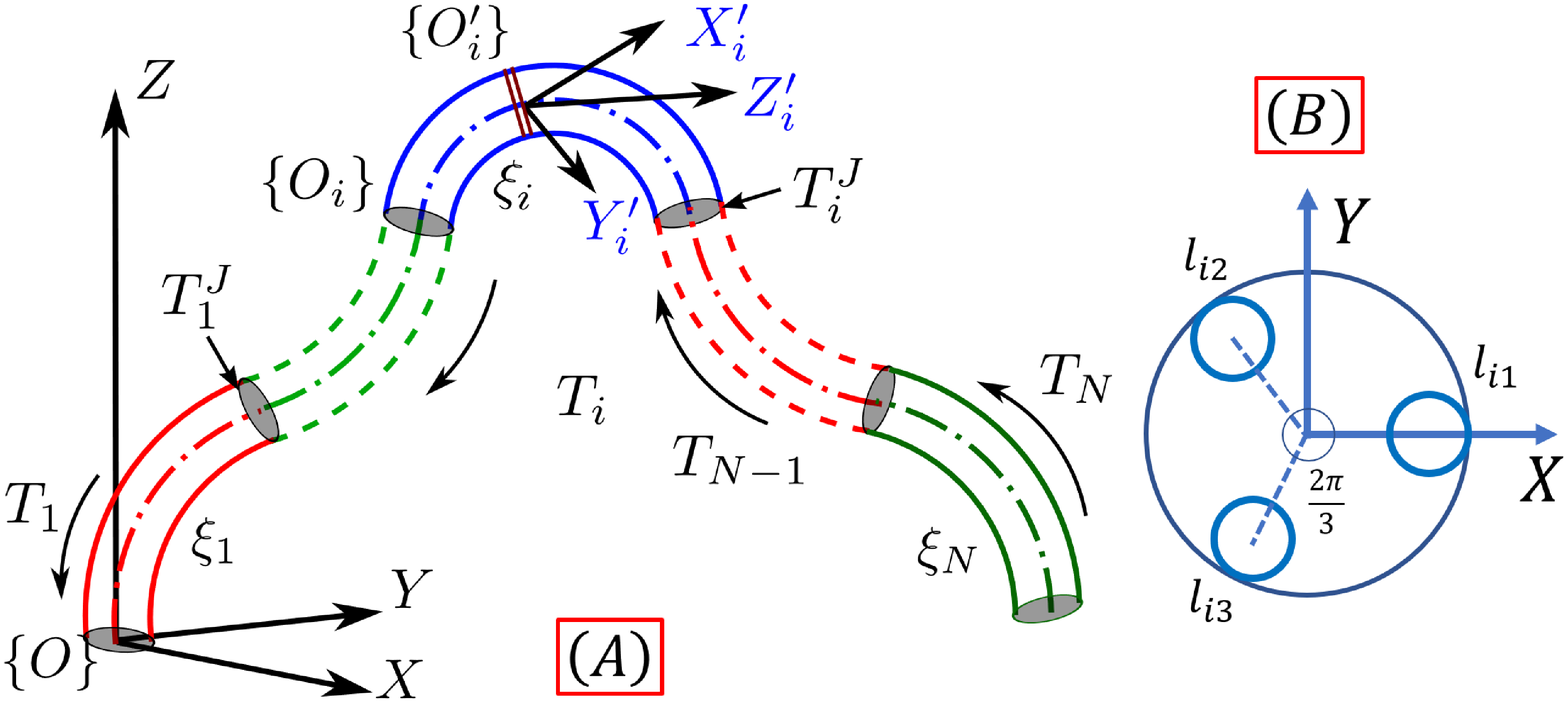}
		\par\end{centering}
	\caption{(A) Schematic of a general $n$-section continuum arm and (B) cross-section of a continuum section showing the actuator arrangement and inextensible backbone in the neutral axis.}
	\label{fig:schematicArm}
\end{figure}

%
%
%
Figure~\ref{fig:schematicArm} shows the schematic of an $n$-section
continuum arm, in which the sections are identical and are enumerated starting
from the base section (index 1) attached to the task-space coordinate
system, $\left\{ O\right\}$. The $i^{th}$ continuum section, shown
in Fig. \ref{fig:schematic}, is actuated by three extending-mode
PMAs that are affixed on rigid terminating plates at either end at
distance $r\in\mathbb{R}^{+}$ from the neutral axis, and that are $\frac{2\pi}{3}$
radians apart. Each continuum section has an inextensible rigid chain
of length $L\in\mathbb{R}^{+}$ in its neutral axis. The unactuated
length of PMAs is $L\in\mathbb{R}^{+}$, and the maximum length change
is $l_{max}\in\mathbb{R}^{+}$. The joint-space vector of the continuum
section is $\boldsymbol{q}_{i}=\left[l_{i1},l_{i2},l_{i3}\right]^{T}$,
where $l_{ij}\in\left[0,l_{max}\right]$ $\forall j\in\left\{ 1,2,3\right\} $.
Individual continuum sections are jointed together using rigid joints
that introduce $\sigma\in\mathbb{R}_{0}^{+}$ linear displacement
along the +Z axis and $\gamma\in\mathbb{R}_{0}$ angular displacement about the
+Z axis of $\left\{ O_{i}\right\} $. 

Along the length of the continuum
section, PMAs are constrained to maintain $r_{i}$ clearance to the
neutral axis. Consequently, any differential length changes of PMAs
due to different pressure inputs cause the continuum section to bend
in a circular arc or extend (when length changes are equal). The subsequent
derivations rely on the assumption that the continuum sections bend
in circular arc shapes. This is a reasonable assumption. As shown in \cite{godage2015modal} and \cite{godage2016dynamics}, under operating conditions where the continuum arms are not subjected to large external forces, continuum sections satisfy this condition. 

\subsection{Derivation of a Reduced DoF Mapping\label{subsec:Derive-a-Reduced}}

The presence of an inextensible backbone introduces an over-constrained
system. Unlike the pneumatically-actuated continuum arms without such a backbone
(see \cite{godage2016dynamics,mcmahan2006field}), for any $i^{th}$ section, the length changes are governed by  

\begin{align}
{\textstyle \sum_{k}}l_{ik} & =0,\label{eq:act_length_constraint}
\end{align}
where physically this means that any extension of actuators results in
the contraction of the others.

As a result, without loss of generality, the complete kinematic model for a continuum section has to be derived from 
the principles of force/moment balance. However, in
this work, by utilizing the above constraint, we will introduce a reduced,
two-DoF model. In this approach, we will disregard
the third actuator, $l_{i3}$, and instead use $l_{i1} 
,l_{i2}$,
and the backbone length to derive a complete mapping. The objective
of our approach is to reduce the overall DoF of the continuum arm, without betraying
the physical behavior, in order to explore the graph-theoretic
path planning approach proposed herein. 

\begin{figure}[tb]
	\begin{centering}
		\includegraphics[width=0.9\columnwidth]{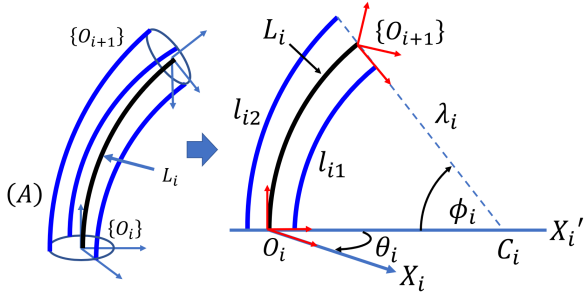}
		\par\end{centering}
	\caption{(A) Schematic diagram showing the actuator and the backbone arrangement
		of the continuum arm; and (B) the schematic of the $i^{th}$ continuum section when looking from an angle normal to the bending plane.}
	\label{fig:schematic}
\end{figure}

By definition, the bending of a continuum section can be described
by three curve parameters, the radius $\lambda_{i}$ of the circular arc,
the angle $\phi_{i}$ subtended by the arc, and the angle $\theta_{i}$ between
the +X axis and the bending plane \cite{godage2015modal}. Figure \ref{fig:schematic}
shows the schematic of the $i^{th}$ section bending in a circular
arc with the arc parameters $\lambda_{i}$, $\phi_{i}$, and $\theta_{i}$.


Utilizing the arc geometry, we can derive the following relations
between the actuator length changes and the curve parameters. A similar
and detailed exposition of the methodology, developed for continuum
arms without constraining backbones, is given in~\cite{godage2015modal}. 

\begin{align}
\begin{split}L_{i} & =\lambda_{i}\phi_{i},\\
L_{i}+l_{i1} & =\left(\lambda_{i}-r_{i}\sin\theta_{i}\right)\phi_{i}, \mbox{and}\\
L_{i}+l_{i2} & =\left(\lambda_{i}-r_{i}\sin\left(\frac{\pi}{3}+\theta_{i}\right)\right)\phi_{i}.
\end{split}
\label{eq:length_relationships}
\end{align}

By manipulating the relations given in \eqref{eq:length_relationships},
we can derive the curve parameters as functions of the length
changes as 

\begin{align}
\begin{split}\theta_{i} & =\arctan\left(l_{i2}\sqrt{3},2l_{i1}-l_{i2}\right),\\
\phi_{i} & =\frac{2\sqrt{l_{i1}^{2}-l_{i1}l_{i2}+l_{i2}^{2}}}{r_{i}\sqrt{3}}, \mbox{and} \\
\lambda_{i} & =\frac{\sqrt{3}L_{i}r_{i}}{2\sqrt{l_{i1}^{2}-l_{i1}l_{i2}+l_{i2}^{2}}}.
\end{split}
\label{eq:curve_parameters}
\end{align}

Using \eqref{eq:curve_parameters}, the homogeneous transformation
matrix (HTM) for the $i^{th}$ section can be derived as

\begin{align}
\mathbf{T}_{i} & =\mathbf{R}_{Z}\left(\theta_{i}\right)\mathbf{P}_{X}\left(\lambda_{i}\right)\mathbf{R}_{Y}\left(\xi_{i}\phi_{i}\right)\mathbf{P}_{X}\left(-\lambda_{i}\right)\mathbf{R}_{Z}\left(\theta_{i}\right)\cdots\nonumber \\
& \qquad\qquad\mathbf{P}_{Z}\left(\sigma_{i}\right)\mathbf{R}_{Z}\left(\gamma_{i}\right)=\left[\begin{array}{cc}
\mathbf{R}_{i} & \boldsymbol{p}_{i}\\
\boldsymbol{0} & 1
\end{array}\right],\label{eq:htm}
\end{align}
where $\mathbf{P}_{X}\in\mathbb{SE}^{3},$ $\mathbf{R}_{Z}\in\mathbb{SO}^{3}$,
and $\mathbf{R}_{Y}\in\mathbb{SO}^{3}$ are HTM that denote translation
along the +X axis, rotation about the +Z and +Y axes, respectively.
$\mathbf{R}_{i}:\left(\boldsymbol{q}_{i},\xi_{i}\right)\mapsto\mathbb{SO}^{3}$
is the resultant rotation matrix and $\boldsymbol{p}_{i}:\left(\boldsymbol{q}_{i},\xi_{i}\right)\mapsto\mathbb{R}^{3}$
is the position vector. The scalar $\xi_{i}$ denotes any point
along the neutral axis, where $\xi_{i}=0$ is the base where $\left\{ O_{i}'\right\} \equiv\left\{ O_{i}\right\} $
and $\xi_{i}=1$ is the tip of the continuum section. We then apply
the $15^{th}$ order multivariate Taylor series expansion on the terms
of \eqref{eq:htm}~to obtain numerically efficient and stable modal
form of the HTM (see \cite{godage2015modal}).

\subsection{Actuator Range Derivation\label{subsec:Actuator-Range}}

In order to preserve the entire \W-space of the continuum section,
the two actuator ranges have 

to be modified. For instance, the actual
PMAs only extend during operation, but as we have omitted $l_{i3}$,
to describe the motion contributions of $l_{i3}$, as per \eqref{eq:act_length_constraint},
we will modify the ranges of $l_{i1}$ and $l_{i2}$. Under the premise that PMA extension is proportional to the differential pressure input, using the resulting curve parameters given by \eqref{eq:curve_parameters} with the length constraint \eqref{eq:act_length_constraint}, one can easily derive the correct ratio of pressures given to PMAs.

\begin{figure}[tb]
	\begin{centering}
		\includegraphics[width=0.7\columnwidth]{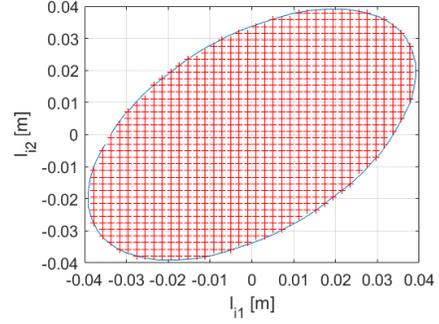}
		\par\end{centering}
	\caption{Actuator range of the reduced mapping.}
	\label{fig:actuator_range}
\end{figure}

Noting the maximum bending, $\phi_{i,max}=\pi$, of the continuum sections,
we can find the valid actuator combinations that result in $\phi_{i}\leq\pi$,
as shown in Fig.~\ref{fig:actuator_range}. We identified this valid length change range by
comparing the actuator combinations in the range $\left[-0.04,0.04\right]$
for $L_{i}=0.15\,m$ and $r_{i}=0.0125\,m$. %
The resulting valid actuator range is give by the rotated ellipse given by 

\begin{align}
g\left(l_{i1},l_{i2}\right) & =al_{i1}^{2}+bl_{i1}l_{i2}+cl_{i2}^{2}+dl_{i1}+el_{i2}+f,\label{eq:ellipse}
\end{align}
where $a=-0.5766$, $b=0.5789$, $c=-0.5766$, $d=0,e=0$, and $f=0.0007$.
This means that if $g(l_{i1}, l_{i2}) \geq 0$ in~\eqref{eq:ellipse}, then the value pair $\left(l_{i1},l_{i2}\right)$ is
a valid combination, and results in $\phi_{i}\leq\pi$. 

\subsection{Kinematic Model\label{subsec:Kinematic-Model}}

Employing the continuum section HTM given in~\eqref{eq:htm} and the
kinematics principles of serial robot chains, the HTM of the $i^{th}$ section
with respect to the \W-space coordinate system $\left\{ O\right\} $,
$\mathbf{T}^{i}:\left(\boldsymbol{q}^{i},\xi_{i}\right)\mapsto\mathbb{SE}^{3}$,
is given by

\begin{align}
\mathbf{T}^{i} & =\prod_{k=1}^{i}\mathbf{T}_{i}=\left[\begin{array}{cc}
\mathbf{R}^{i} & \boldsymbol{p}^{i}\\
\boldsymbol{0} & 1
\end{array}\right].\label{eq:HTM_ith}
\end{align}

The HTM in \eqref{eq:HTM_ith} can be expanded to obtain the recursive
form of the kinematics as

\begin{align}
\begin{split}\mathbf{R}^{i} & =\mathbf{R}^{i-1}\mathbf{R}_{i},\\
\boldsymbol{p}^{i} & =\boldsymbol{p}^{i-1}+\mathbf{R}^{i-1}\boldsymbol{p}_{i},
\end{split}
\label{eq:pR_recursive}
\end{align}
where $\mathbf{R}^{i-1}:\left(\boldsymbol{q}^{i-1}\right)\mapsto\mathbb{SO}^{3}$
and $\boldsymbol{p}_{i}:\left(\boldsymbol{q}^{i-1}\right)\mapsto\mathbb{R}^{3}$
is the section tip rotation matrix and position vector of the preceding
continuum section, respectively. Notice the absence of $\xi_{i}$, as $\xi_{k}=1, \forall k<i,$
by the definition of $\xi_{i}$ (see \cite{godage2015modal}).

\section{The Algorithm}\label{sec:method}
We assume familiarity with basic graph algorithms and terminologies, and refer the reader to the textbook~\cite{clrs}.

The approach we employ starts by constructing a mapping between the \C-space and the \W-space of the continuum arm. This is done by discretizing the pneumatic pressures of the three arm-sections to an appropriate granularity level that allows a feasible enumeration of the corresponding \C-space configurations, without trading off (by much) the quality of the path planning achieved. After enumerating the \C-space configurations, we construct an auxiliary graph of the \C-space. Path planning is then performed on this auxiliary graph and mapped back to the \W-space. We proceed to the details.

\subsection{The Auxiliary Graph}\label{sub:auxG}

We construct an auxiliary graph, $G_C$, whose vertices correspond to the enumerated configurations, and whose edges correspond to adjacent configurations (\ie, configurations that are obtained from one another via a one-step variation in a single DoF). Since one of our goals is achieving smooth path planning, we associate weights with the edges of the auxiliary graph that reflect the \emph{change in the orientation} between the two configurations that are endpoints of the same edge. This is quantified by measuring the distance between two vectors whose coordinates are formed by the six angles $\theta_i$ and $\phi_i$ (see Fig.~\ref{fig:schematic}), corresponding to the three arm sections (\ie, two angles per section).

\subsection{Path Planning}\label{sub:pplan}
In the path planning problems under consideration, we are given two points, a starting point $s$ and  a destination $t$, in a 3-dimensional \W-space, and we wish to compute a smooth $s$-$t$ path for the continuum arm. We considered both cases: where the \W-space is obstacle-free and where it contains obstacles. 

To perform path planning, we start by discretizing the \W-space by creating a cubic-grid, in which each cube has the same dimension (1 unit). Given two points $s$ and $t$ in the 3-dimensional \W-space that correspond to two \C-space configurations, between which we wish to compute a (smooth) path, we first determine the two cubes $Q_s$ and $Q_t$ containing $s$ and $t$, respectively. We then find a shortest---w.r.t.~the Euclidean distance---{\em cube path} between $Q_s$ and $Q_t$, which is a sequence of adjacent cubes in the cubic-grid that starts at $Q_s$ and ends at $Q_t$. This step is done by constructing a graph whose vertices are the cubes in the grid, and whose edges correspond to adjacent cubes; we refer to this graph as the \emph{cubes-graph}, and denote it by $G_Q$. An edge in $G_Q$ has weight equal to the Euclidean distance between the centers of the two cubes corresponding to the endpoints of the edge. We then find a shortest path $P_{st}$ in $G_Q$ between $Q_s$ and $Q_t$; this can be done by applying Dijkstra's algorithm. Let $P_{st}=(Q_0=Q_s, Q_1, Q_2, \ldots, Q_r=Q_t)$. Based on the enumeration of the $C$-space, we can find the set of configurations, $C_i$, whose corresponding points in the \W-space fall within cube $Q_i$, for $i=1, \ldots, r$; we let $C_0$ be the singleton containing the configuration corresponding to the initial point $s$. In the case where obstacles are present, we purge from each $C_i$ the configurations that are invalid because they correspond to a position in which the continuum arm intersects an obstacle in the \W-space. In order to test that, for each configuration, we keep track of intermediate points (in the \W-pace) along the three sections of the continuum arm, and use those points to detect intersection with obstacles.

The goal now becomes to compute a shortest path in $G_C$ that starts at $C_0$, proceeds to visit a vertex in each $C_i$, for $i=1, \ldots, r-1$ (in the listed order), and ends up at a vertex in $C_r$. To compute such a path, and in order to reduce memory storage (as storing the whole configurations graph $G_C$ turns out to be quite inefficient) and exploit parallelism, we employ a shortest-path algorithm that takes advantage of the structure of the layered graph, $G_L$, whose vertex-set consists of the configurations in $\bigcup_{i \in \{0, \ldots, r\}}C_i$, and edge-set consists of those edges in $G_C$ that join configurations in consecutive layers $C_i, C_{i+1}$, for $i \in \{0, \ldots, r-1\}$. For this purpose, we apply a variant of the Bellman-Ford's algorithm (see~\cite{clrs} for Bellman-Ford's algorithm) for computing a shortest path between the single configuration in $C_0$ and every vertex in $C_r$. Here, We do not insist on reaching $t$, and settle for any configuration corresponding to a point in $t$'s cube, and hence is close enough to $t$, \ie, within distance $\sqrt{3}\cdot d$ from $t$, where $d$ is the cube dimension. 
The algorithm starts by initializing the cost for $C_0$ to $0$, and for each other configuration in $G_L$ to $\infty$, which is a sentinel value that is assumed to be larger than any concrete number. The algorithm employs the subroutine {\bf Relax($u$, $v$)}, which operates on edge $uv$ and \emph{relaxes} $uv$ by updating $cost(v)$ to become $cost(u) + wt(u, v)$, in case the current value of $cost(v)$ exceeds $cost(u) + wt(u, v)$. The algorithm then performs $r$ iterations, where in iteration $i$, $i =0, \ldots, r-1$, the algorithm relaxes all edges between layer $C_i$ and layer $C_{i+1}$ in $G_L$. (This computation is parallelized by running it separately on the set of all edges incident to the same node in $C_{i+1}$. Moreover, after iteration $i$, the vertices in $C_i$ are no longer useful during the rest of the algorithm, and hence, can be removed from memory and replaced with those in $C_{i+2}$.) At the end of step $r-1$, we have computed the shortest path between the single configuration in $C_0$ and every vertex in $C_r$. The shortest path among among all those (shortest paths) is then mapped to an $s$-$t$ path in the \W-space.  

\section{IMPLEMENTATION AND RESULTS}\label{sec:results}
We implemented and tested our algorithm on a Windows-10 computer, with 64-bit CPU, 6~core 3.20~GHz processor, and 64~GB RAM. Several considerations were made over the course of the project to reduce the CPU and memory load. First of all, most of the calculations were performed using the ``NumPy'' Python library \cite{developers2013numpy}. The array and matrix operations in ``NumPy'' are more efficient than the standard library, as they are optimized to use less memory and CPU time. We encoded the cube indices and the coordinates of the points as decimal numbers to save memory. In order to determine the points in the \W-space, given their corresponding configurations, we relied on large matrix multiplication. To speed up this process, two features were used. First of all, the matrix is modified so that it can be vectorized; this has the effect of parallelization without any explicit threads, and  improves performance. The other feature we used is the ``JIT" compiler of the Numba Library \cite{lam2015numba} to improve performance. (For instance, the function that calculates the shortest distance between points on the path is aggressively vectorized.) These techniques brought the time taken to generate the \C-space graph down from 16 hours to 10 minutes. To avoid recalculating the graph each time, we used the NumPy save function and the ``pickle'' python library\footnote{https://docs.python.org/3/library/pickle.html} to store the graph, cube list, and point list on disk for easy reuse. Lastly, in order to perform graph operations, we used the python library ``igraph''\cite{csardi2010igraph}. This is a high-performance library that handles all of the graph generation and simple path planning for the cubes. 

To test our algorithm, we discretized the three sections of the continuum arm; each section has 2 DoF, and each DoF was uniformly discretized into (roughly) 26 steps. This resulted in sampling 688 configurations for the tip of each section, and $688^3=325,660,672$ configurations for the tip of the whole arm. Thus, the number of vertices in the \C-space graph is $325,660,672$. Based on the points in the \W-space corresponding to the sampled configurations, we placed a bounding box of $X$-$Y$-$Z$ dimensions $82\,\text{cm}\times 82\,\text{cm} \times 74\,\text{cm}$ that contains all these points. This bounding box was discretized into cubes, each of dimension 1~cm. 

We ran 2,000 tests, 1000 tests in which the \W-space is obstacle-free, and 1000 tests in which the \W-space contains either 2 or 3 spherical-shape obstacles.
The starting and ending points of the path were generated randomly from the \W-space (by randomly choosing two configurations), and so was the center of the spherical-shape obstacles; the radius of the obstacles was randomly chosen from the interval $[3\,\text{cm}, 25\,\text{cm}]$
. We then located the two cubes containing the starting and ending points; computed a shortest cube-path between these two cubes in the cubes graph; and then ran our algorithm on the corresponding layered graph resulting from the cubes on this cube-path, as explained in the previous section.

The IK approach was run on the same generated test cases. Table~\ref{tab:1} below summarizes the results obtained for both approaches. The third column of the table shows the success rate of the IK approach and the fourth column shows the average time (in seconds) taken by the IK approach (over all cases tested). Columns 5 and 6 show the corresponding data for our algorithm (Alg), given in the previous section. 
%

\begin{table}[tb]
	\caption[]{Summary of the numerical results.}
	\label{tab:1}
	\centering{}%
	\begin{tabular}{@{}l@{\quad}|l@{~~~}l@{~~~}l@{~~~}l@{~~~}l@{}} 
		
		Scenario & \# Cases & I.~K. Rate & I.~K. Time & Alg.~Rate & Alg.~Time \\
		\hline
		No obstacles &	1000	& 74\%	& 0.01 & 	100\%	& 75.43 \\   
		Obstacles	& 1000	& 68.67\%	& 0.45	& 100\%	& 85.55
	\end{tabular}
	
	
\end{table}

\begin{figure}[tb]
	\begin{centering}
		\includegraphics[width=1.0\columnwidth]{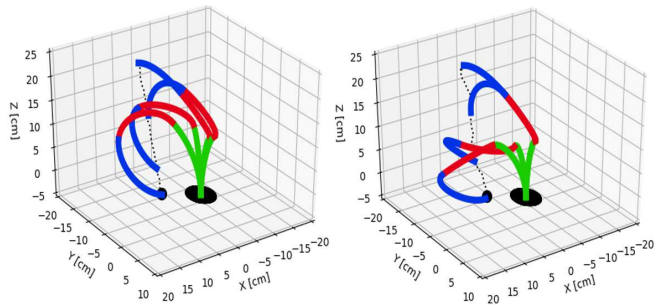}
		\par\end{centering}
	\caption{Illustrative test case where both the proposed approach (Right) and the IK (Left) successfully found solutions. No obstacles were present in the work-space.}
	\label{fig:noObs_success}
\end{figure}

\begin{figure}[tb]
	\begin{centering}
		\includegraphics[width=1.0\columnwidth]{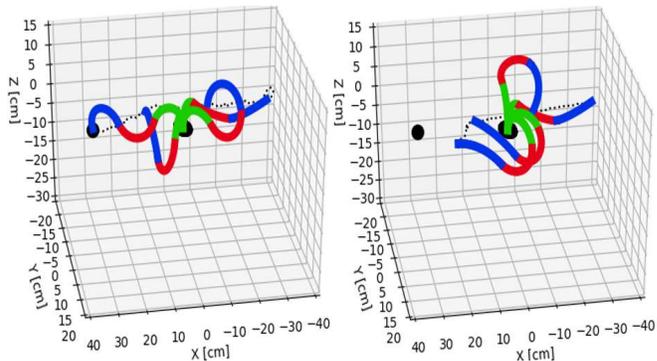}
		\par\end{centering}
	\caption{An example test case where the proposed approach (Right) converged to a solution, but the IK (Left) was unsuccessful. No obstacles were present in the work-space.}
	\label{fig:noObs_oursuccess}
\end{figure}

Figure~\ref{fig:noObs_success} shows a test case in an obstacle-free space where both the proposed and the IK approaches succeeded in finding a solution. Given the (continuous) nature of the IK approach, the resulting path generated by the IK approach is shorter and smoother than the one computed by the proposed approach. (Our approach is limited by the rough discretizations of the \C-Space and \W-Space in order to ensure a feasible computation.) The reliability of our approach stems from the fact that it has apriori access to the entire (discretized) \C-space, which it could use to avoid ``knotting" or running into local minima. As shown in Fig.~\ref{fig:noObs_oursuccess}, in this example our planner was able to find a 
path
, whereas the IK approach did not even converge to a solution. It can be seen that the base and the mid sections of the continuum arm form a Knot~\cite{xiao2010real}, as the IK approach tries to minimize the distance to the target point but 
the mid and base sections are already at the maximum bending. 
Consequently, the IK approach failed to yield a solution. This is the common reason behind the failure of IK-based continuum arm path planners. Since the \C-space graph---constructed according to a proper discretization---allows us to capture a wide range of motion for the continuum arm covering the entire task-space, our algorithm has the ability to adjust itself, when started from certain initial configurations, to reach the destination point, in situations where the IK approach, starting from the same initial configurations, is incapable of doing so, and diverges to points that are far away from the destination points.

\begin{figure}[tb]
	\begin{centering}
		\includegraphics[width=1.0\columnwidth]{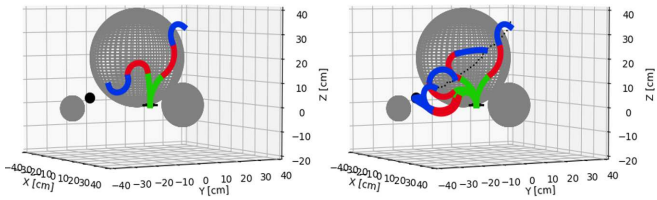}
		\par\end{centering}
	\caption{One scenario in which the proposed planner (top) was able to find a path from the starting point to the end point amidst obstacles whereas the IK (bottom) planner failed.}
	\label{fig:Obs_oursuccess}
\end{figure}

Figure~\ref{fig:Obs_oursuccess} shows a test case with obstacles. Due to the highly-constrained nature of the problem, unlike the obstacle-free scenario, in this case it is not possible to generate a meaningful shortest path between the starting and end point. Therefore, in favor of a fair comparison with the IK approach, we removed the shortest path requirement so that the IK planner is only required to find \emph{a} path. In the IK planner, we formulated the obstacle avoidance as a constrained function; we considered ten points along a continuum section (30 points total), and the constraint function must maintain a positive distance between all the points and obstacles. In contrast, the proposed planner solves for the shortest path.  As shown in Fig.~\ref{fig:Obs_oursuccess}, in this test, the IK failed to converge. Similarly to the obstacle-free test case shown in Fig.~\ref{fig:noObs_oursuccess}, the IK planner was stuck in a local minima (knot configuration), whereas, due to the apriori knowledge of the \W-Space, our planner was able to find poses that circumvent configurations associated with local minima. 

Since we generate the obstacles randomly, there is the possibility of having cases where there is no solution; for instance, this is the case if the obstacles form a separator in the \W-Space, disconnecting the starting point from the end point.  We discarded such cases when computing the success rate in Table~\ref{tab:1}, and computed the success rate of IK as a percentage of the successful cases, \ie, in which a path exists. (Note that, given how our planner works, it is able to find a solution if there is a path.) Consequently, the proposed approach is near-optimal and reliable for planning path of multisection continuum arms in spaces with or without obstacles. In addition, it facilitates the introduction of multiple optimization criteria---such as smooth trajectories, in contrast to the IK approaches, which will increasingly fail as the number of constraints grow. In future work, we will explore the possibility of using other optimality criteria, such as energy efficiency and dynamic stability, for path planning.

\section{CONCLUSIONS}\label{sec:conclusion}
Continuum arms have seen a surge of research in recent years. However, due to the complex and nonlinear kinematics associated with continuum structures, limited research has been conducted on path planning. Most of such planners have been based on inverse kinematics, and therefore resulted in poor reliability due to their potential of running into local minima. This paper presented a near-optimal smooth path planning approach for multisection continuum arms using a graph-theoretic approach, which is based on a high-fidelity mapping between the configuration space and the work space of the arm.  The proposed approach was then tested against the classical IK solvers in work spaces with and without obstacles. The proposed approach was able to solve the path planning problems (100\% without obstacles) where the IK consistently reported poor reliability (70\% without obstacles). For tests with randomly placed obstacles in the work space, the proposed planner reported 100\% success rate where there exists a solution, while the IK yielded 69\%
success rate.


\bibliographystyle{IEEEtran}
\bibliography{refs}

\end{document}